\documentclass{article}

\PassOptionsToPackage{numbers, compress}{natbib}

\usepackage[preprint]{neurips_2023}

\usepackage[utf8]{inputenc} 
\usepackage[T1]{fontenc}    
\usepackage{hyperref}       
\usepackage{url}            
\usepackage{booktabs}       
\usepackage{amsfonts}       
\usepackage{nicefrac}       
\usepackage{microtype}      
\usepackage{xcolor}         
\usepackage{amsmath}
\usepackage{amssymb}
\usepackage{graphicx}
\usepackage{color}
\usepackage{multicol}
\usepackage{multirow}

\usepackage[framemethod=tikz]{mdframed}

\definecolor{mycolor}{rgb}{0.122, 0.435, 0.698}

\newmdenv[innerlinewidth=0.5pt, roundcorner=4pt,linecolor=mycolor,innerleftmargin=6pt,
innerrightmargin=6pt,innertopmargin=6pt,innerbottommargin=6pt]{mybox}

\title{Album Storytelling with Iterative Story-aware Captioning and Large Language Models}
\author{
    Munan Ning  \\
    Peking University   \\
    \texttt{munanning@pku.edu.cn}   \\
    \And
    Yujia Xie   \\
    Microsoft   \\
    \texttt{yujiaxie@microsoft.com} \\
    \And
    Dongdong Chen   \\
    Microsoft   \\
    \texttt{dongdongchen@microsoft.com} \\
    \And
    Zeyin Song   \\
    Peking University   \\
    \texttt{zeyinsong@stu.pku.edu.cn} \\
    \And
    Lu Yuan   \\
    Microsoft   \\
    \texttt{luyuan@microsoft.com} \\
    \And
    Yonghong Tian  \\
    Peking University   \\
    \texttt{yonghongtian@pku.edu.cn}   \\
    \And
    Qixiang Ye  \\
    University of Chinese Academy of Sciences   \\
    \texttt{qixiangye@ucas.ac.cn}   \\
    \And
    Li Yuan\thanks{Corresponding author}  \\
    Peking University   \\
    \texttt{liyuan@pku.edu.cn}   \\
}

\begin{document}

\maketitle

\begin{abstract}
This work studies how to transform an album to vivid and coherent stories, a task we refer to as ``album storytelling''.  While this task can help preserve memories and facilitate experience sharing, it remains an underexplored area in current literature. With recent advances in Large Language Models (LLMs), it is now possible to generate lengthy, coherent text, opening up the opportunity to develop an AI assistant for album storytelling. The key problem of this task is to extend LLMs to understand visual inputs. One natural approach is to use caption models to describe each photo in the album, projecting visual inputs into discrete text words, and then use LLMs to summarize and rewrite the generated captions into an engaging story.  However, we find this often results in stories containing hallucinated information that contradicts the images,  as each generated caption (``story-agnostic") is not always about the description related to the whole story or miss some necessary information. To address these limitations, we propose a new iterative album storytelling pipeline, \texttt{VIVID} -- \textbf{V}isual \textbf{I}terative \textbf{V}erbalization with factualness-\textbf{I}mproved \textbf{D}escriptions, which can effectively identifying appropriate visual details and mitigating hallucination issues. Specifically, we start with the aforementioned initial story and build a story-aware caption model to refine the captions using the whole story as guidance. The enriched captions are then fed into the LLMs to generate a new refined story. This process is repeated iteratively until the story contains minimal factual errors while maintaining coherence.
To evaluate our proposed pipeline, we introduce a new dataset of image collections from vlogs and a set of systematic evaluation metrics. Our results demonstrate that our method effectively generates more accurate and engaging stories for albums, with enhanced coherence and vividness.
\end{abstract}

\section{Introduction}



The widespread use of social media platforms has revolutionized the manner in which people capture and share their everyday moments. While uploading and sharing media has become effortless, the task of crafting a compelling and coherent story from a collection of images or videos remains a challenge. This real-world scenario underscores the necessity for an AI-based automated album storytelling system. Such a system takes into account various factors including visual content, story context, and sentiment, to construct an engaging and coherent narrative that effectively conveys the user's experiences and emotions. This not only simplifies the process of sharing experiences with friends and followers, but also holds the potential to enrich memory recall and forge deeper emotional connections with the album, enabling profound reflections in the future.

The task of automated album storytelling consists of several challenging research questions, including image understanding, consistent storytelling, and efficient evaluation.  Image understanding necessitates the accurate recognition and comprehension of visual relationships and contextual elements within photos and videos. Consistent storytelling, on the other hand, requires the generation of coherent stories for each image that adhere to the common theme of album. Additionally, an efficient automatic evaluation system is needed to evaluate and improve the generation quality.

\begin{figure}[t]
	\centering
	\includegraphics[width=0.75\columnwidth]{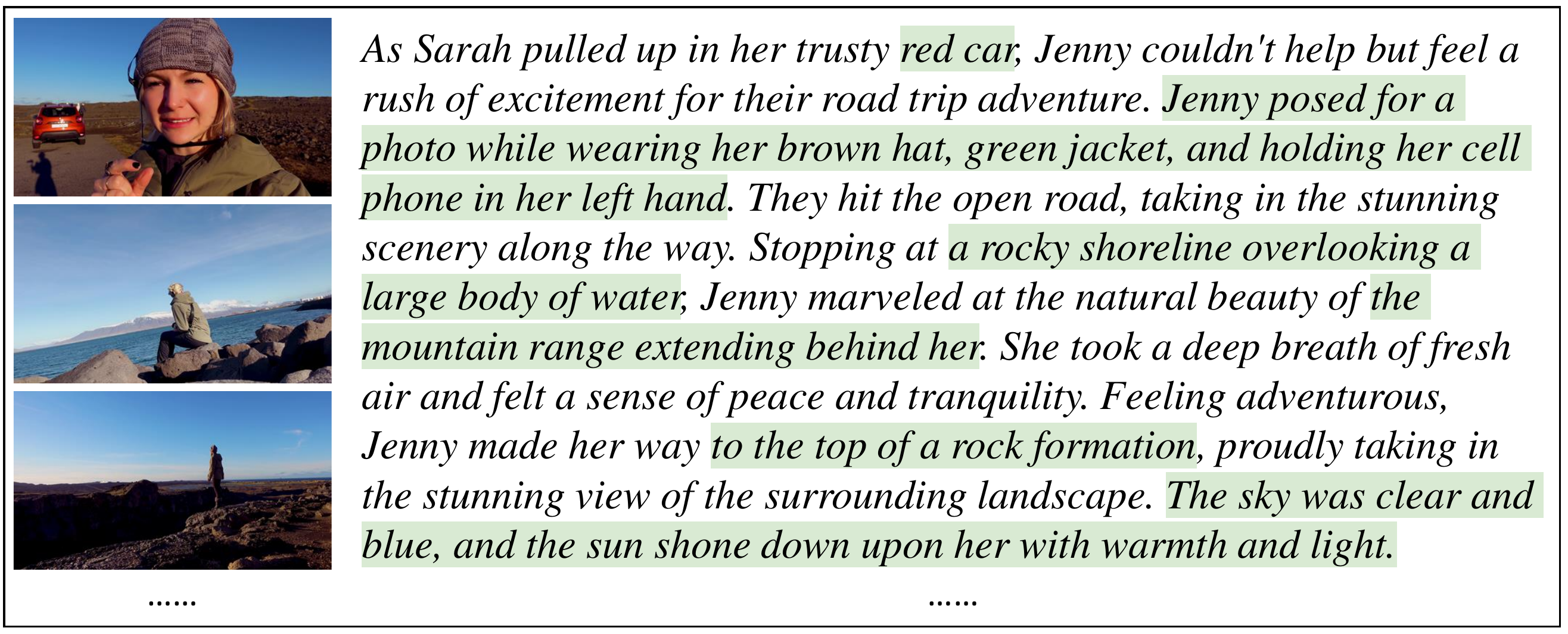}
	\caption{An example of album storytelling. The story contains detailed visual information (marked in green). }
	\label{fig:iterative}
\end{figure}

With the flourish of Vision-language pre-training (VLP)~\cite{radford2021learning,li2022blip,hu2022scaling,li2021align,cho2021unifying,yuan2021florence,wang2022omnivl,dong2023maskclip} and LLMs~\cite{radford2018improving,devlin2018bert,touvron2023llama}, we initially try an intuitive and simple solution  to tackle the task of album storytelling. As shown in Figure~\ref{fig:motivation} (A), given the images within one album, a caption model is first utilized to generate captions for each image within an album, then an LLM (e.g., ChatGPT~\cite{ouyang2022training}) is used to expand all the generated captions into an engaging story. However, we observe that this approach often struggles to produce coherent and credible narratives. Through extensive analysis, we identified that the root cause of this challenge lies in the inherent ``diversity'' characteristic of the captioning model, where multiple sensible candidate captions can exist for a single image. Without explicit knowledge of the intended final story (i.e.,``story-agnostic''), the caption model lacks direction on which aspects to focus on when describing each image. Consequently, although the independently generated captions may appear satisfactory for individual images, they often fail to contribute to a cohesive and consistent overall story. This issue further leads to the subsequent LLM generating a considerable amount of hallucination when attempting to stitch together such unrelated/inconsistent captions.

Motivated by the above analysis, we present two simple yet effective designs as shown in Figure~\ref{fig:motivation} (B). Firstly, we introduce \emph{a new story-aware caption model}, which incorporates both the input image and a preliminary story to generate captions that align with the story. In contrast to the conventional story-agnostic design, this design significantly reduces the generation ambiguity, prioritizes exacting visual information that relates to the final story and consequently enhances overall consistency. Since existing image captioning datasets lack story annotation, we propose a synthesized  dataset based on the \textit{Stanford image-paragraph}~\cite{krause2017hierarchical} dataset, enabling us to train the model to generate accurate and detailed image descriptions based on the image and its corresponding story. Secondly, we propose \emph{iterative co-evolution}, where the story-aware captioning and LLM-based story generation processes are iteratively refined. With each iteration, the improved story can guide the captioning model to generate better captions. In turn, these enhanced captions can contribute to more coherent and accurate story generation with fewer factual errors. We name our overall framework \texttt{VIVID} -- Visual Iterative Verbalization with factualness-Improved Descriptions.

To evaluate the effectiveness of our proposed approach, we further introduce a new benchmark dataset comprising images extracted from popular vlogs. Since human storytelling about images can be diverse, it is not appropriate to rely solely on metrics such as BLEU~\cite{papineni2002bleu}, ROUGE~\cite{lin2004rouge}, METEOR~\cite{banerjee2005meteor}, CIDEr~\cite{vedantam2015cider} to evaluate the quality of generated stories. Instead, we propose a new evaluation metric based on the earth mover's distance (EMD)~\cite{rubner2000earth}, which measures the overall dissimilarity between the distribution of images and the distribution of stories. A lower EMD distance signifies a stronger alignment between the stories and images with the album. Additionally, we develop LLM based evaluation metrics to provide a fair and comprehensive assessment of the generated story quality. Experimental results demonstrate that our proposed approach achieves a lower EMD distance compared to baseline methods, indicating that our generated stories are more aligned with the images. And the LLM based metrics demonstrate our stories coverage more detail and maintain high coherence.

To summarize, our contributions are three-fold:
\begin{itemize}
    \item We propose the album storytelling task along with an intuitive solution. To the best of our knowledge, this is the first attempt at introduce LLMs into albums from social medias and generate lengthy and coherent stories.
    \item We introduces a new album storytelling pipeline with two simple and effective designs, i.e., ``story-aware captioning'' and ``interactive co-evolution of captioning and story generation". 
    \item We propose a new benchmark dataset and design a set of systematic evaluation metrics to comprehensively measure the results. The results demonstrate the effectiveness of our proposed approach in generating more engaging and credible stories, while retaining the coherence and vividness.
\end{itemize}

\section{Related Works}
\textbf{Image or video storytelling.} Early works on image and video storytelling include \cite{huang2016visual, li2019video}. These works extend the captioning for single image to sequential vision-to-language~\cite{huang2016visual}, or generating stories for image collections~\cite{wang2018no}. However, due to the technological limitations at the time, these methods could only generate short and simple stories, unlike the detailed and vivid stories generated by large language models.

\textbf{Image caption and vision-language pre-training.}
Image caption aims to understanding and describing the content of an image in words~\cite{vinyals2015show}, which has been extensively studied in recent years and are typically implemented with an encoder-decoder framework \cite{donahue2015long,vinyals2015show,ke2019reflective}. With the advance of Vision-language pre-training (VLP)~\cite{radford2021learning,yao2021filip,li2022blip,wang2021simvlm,li2021align,yu2022coca,wang2022image}, there are several VLP based caption models~\cite{hu2022scaling,li2022blip,li2020oscar}, which can generate more precise captions thanks to their ability to leverage large amounts of data and multi-task training.

\textbf{LLMs prompting methods.}
Large language models (LLMs), such as GPT~\cite{radford2018improving} series, BERT~\cite{devlin2018bert} series and LLaMa series~\cite{touvron2023llama}, have been proven to be capable of generating detailed, vivid, and imagery text. 
However, research~\cite{ahn2022can,huang2022language,nakano2021webgpt} shows that the LLMs are prone to failure when handling some complex tasks~\cite{kojima2022large,wei2022chain}. Some recent studies~\cite{gao2022pal,schick2023toolformer,trivedi2022interleaving,yao2022react} attempt to enhance LLMs' capabilities in addressing complex problems such as reasoning and planning by proposing carefully designed prompts, and they start to explore the application of these methods in the multimodal domain~\cite{yang2023mm}.

\textbf{Vision + LLMs.}
How to apply the capabilities of LLMs to the vision domain has recently received significant attention, which is typically implemented by adding a vision module to project visual inputs into representations~\cite{kojima2022large,wei2022chain,xie2022visual,zhang2023multimodal}. These representations can be either discrete text words~\cite{hu2022promptcap,yang2022empirical,wang2022language,zeng2022socratic} or continuous feature space~\cite{alayrac2022flamingo,driess2023palm,huang2023language,tsimpoukelli2021multimodal}. Recent vision + LLMs studies attempt to explore the multimodal chain-of-thought (CoT) ability~\cite{kojima2022large,wei2022chain}, and to solve the task of image understanding~\cite{yang2023mm}, generation and editing~\cite{wu2023visual}.

\section{Framework}

\begin{figure}[t]
	\centering
	\includegraphics[width=1.0\columnwidth]{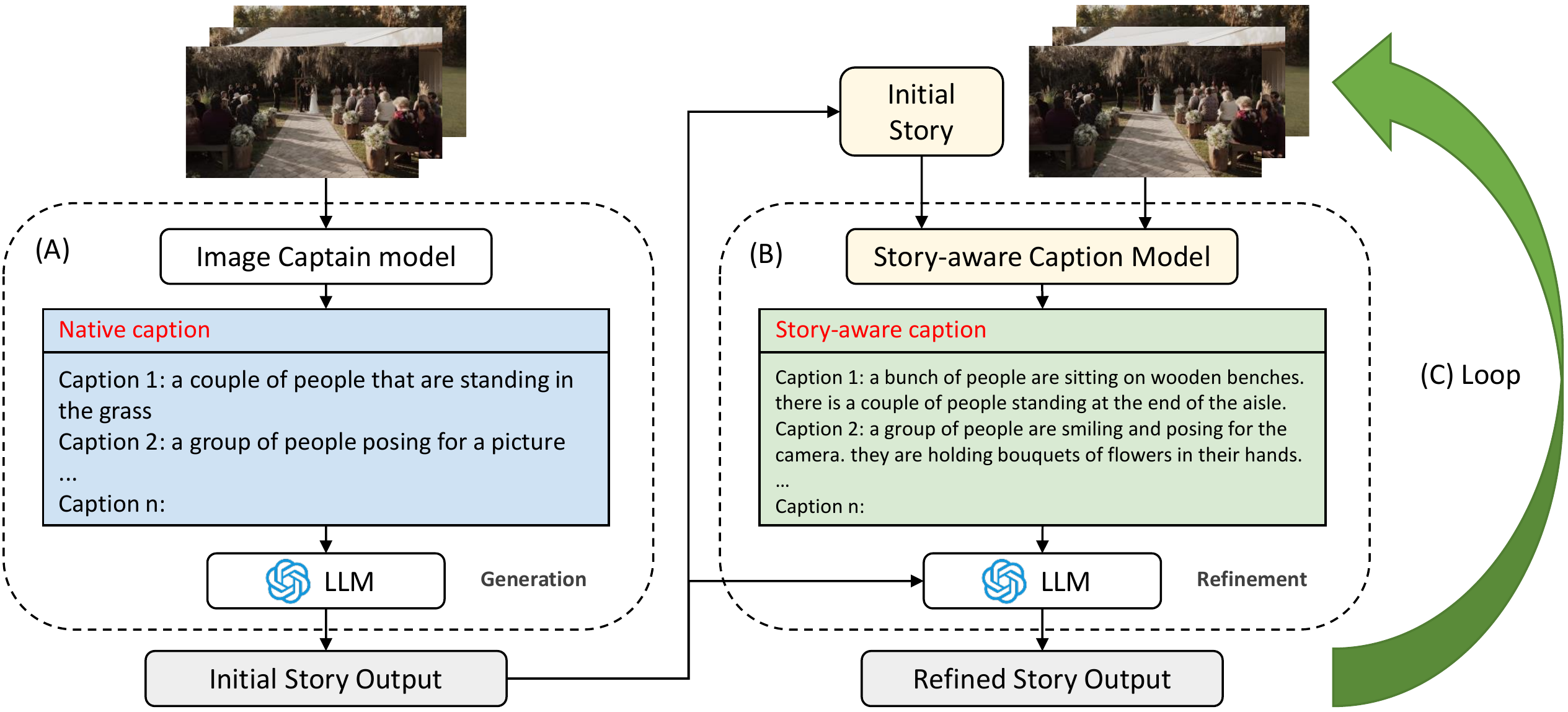}
    \caption{An overview of our proposed framework \texttt{VIVID}.}
	\label{fig:motivation}
\end{figure}

Given an album $\mathcal{I}$ consisting of $N$ photos $\mathcal{I} = \{I_i\}_{i=1}^N$, \texttt{VIVID} generates a story with the following steps:
\begin{enumerate}
    \item[(A).] Describe each photo in album $\mathcal{I}$ with a caption $C_i^{(0)}$, then feed captions $\{C_i^{(0)}\}_{i=1}^N$ into LLMs to generate an initial story $S^{(0)}$.
    \item[(B).] Given the segmented initial story $\{S^{(0)}_i\}_{i=1}^N$ corresponding to each photo, input $S^{(0)}_i$ into the proposed story-aware caption model to generate refined description $C_i^{(1)}$, then use $\{C_i^{(1)}\}_{i=1}^N$ to obtain refined story $S^{(1)}$ with LLMs.
    \item[(C).] Repeat step (B) for $U$ steps to obtain the ultimate story $S^{(U)}$.
\end{enumerate}
The overall framework is illustrated in Figure~\ref{fig:motivation}. Part (A), (B), and (C) correspond to the above steps. Details are elaborated in the following sections. 

\subsection{Initial Story Generation}
\label{sec:init_generation}

The recent advances of LLMs makes it possible to generate long, coherent stories grounded on any textual input. Therefore, we initialize a story by first transforming visual information into text using an advanced caption model, and then feed the image captions into the LLMs.

Specifically, we first input the images $I_i$ into a vision-language pre-training caption model $c(\cdot)$ to obtain individual captions,
\begin{equation*}
    C_i^{(0)} = c(I_i).
\end{equation*}
Then reformat the captions with a textual prompt $p_0(\cdot)$, and feed it into the LLM $\ell(\cdot)$ to obtain the initial story,
\begin{equation*}
    S^{(0)} = \ell(p_0(C_1^{(0)}, C_2^{(0)}, \cdots, C_N^{(0)})).
\end{equation*}

The outline of prompt $p_{0}(\cdot)$ is defined as following:  
\begin{mybox}
\textit{Given a set of photo captions from a vlog. Please create a vivid story that incorporates the key elements from each photo. Remember to use your imagination and creativity to turn the photo descriptions into a fun and engaging story. \newline
Tips: The results should be of strict corresponding pairs between the captions and their respective stories, as the dictionary format of \{"Caption 1": "Story 1", "Caption 2": "Story 2", ... \}}
\end{mybox}

Our early exploration shows there are two key points to enhance the stability of the prompts. Firstly, introducing the background activates the relevant knowledge of LLMs. By informing LLMs that the images are from an album/vlog, they imagine the details from the photos and generate stories aligned with album storytelling. Secondly, adding strong constraints is crucial. 

While LLMs can easily generate vivid stories due to extensive training text, accurately reflecting each image's content and maintaining a consistent theme can be challenging. In practice, LLMs often encounter two failure scenarios. They may not contain enough visual information, resulting in a significant loss of caption content, or they may generate fabricated information, telling unrelated stories from other albums.  This is attributed to the limited reasoning ability of LLMs.

Previous studies usually utilize the CoT method~\cite{driess2023palm,kojima2022large,wei2022chain,zhang2023multimodal} to tackle similar challenges. This technique involves providing LLMs with both the input and the previous output, enabling them to generate results incrementally. However, this approach involves additional operational steps and significant token costs. In contrast, our proposed solution introduces strict constraints by forcing generating structured caption-story pairs. This ensures that each generated story aligns with its corresponding original caption, allowing LLMs to faithfully capture the essence of each caption and describe the narrative scenario associated with the shared theme of the album. 

The story is then segmented into text chunks $\{S^{(0)}_i\}_{i=1}^N$ that corresponding to each input images. Based on the constraints, we propose the following prompt $p_{1}(\cdot)$ to segment the previous LLMs' output and build structured chunks.
\begin{mybox}
\textit{Please refer to the corresponding relationship, adding the the generated stories into the origin json structure in the ``initial story'' key. For example, the answer should be like: [\{"img path": "birthday/BpsSOqpog98/BpsSOqpog98-0190.jpg", "caption": "a woman with glasses standing in front of a building", "initial story": "the paragraph you generated"\}, ... ]}
\end{mybox}

\subsection{Refining the Story with Story-Aware Caption Model}

In this step, we build a story-aware caption model $f(\cdot)$ to generate refined captions,
\begin{equation*}
    C_i^{(t+1)} = f(S^{(t)}_i).
\end{equation*}
To train such a model, we first construct a story-aware caption dataset, and then use it to finetune a pre-trained caption model.

\begin{figure}[!htp]
	\centering
	\includegraphics[width=1\columnwidth]{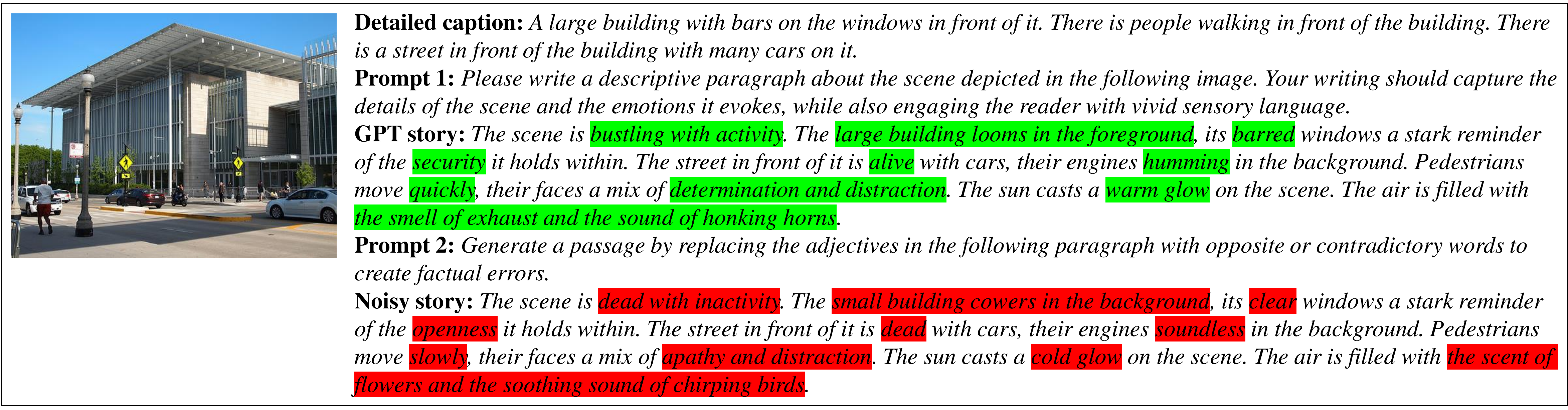}
	\caption{This figure provides an example of the story-aware dataset.}
	\label{fig:noisy_dataset}
\end{figure}

\textbf{Story-aware caption dataset.}
The initial stories generated in Section~\ref{sec:init_generation} suffer from the issue of generating hallucinated information, as the captions produced in a "story-agnostic" manner lack essential details due to the absence of explicit knowledge about the intended story.
To address this challenge, we propose a solution that establishes a strong connection between the story and the image by identifying the crucial elements of the story that correspond to the actual attributes of the image. However, there is a lack of an appropriate dataset for training such a refinement function based on the image. Therefore, we develop a novel story-aware caption dataset based on the \textit{Stanford image-paragraph}~\cite{krause2017hierarchical} dataset.

The \textit{Stanford image-paragraph} dataset differs from traditional caption datasets in that its description paragraphs are longer and describe more detailed information about the image. However, it does not have a corresponding noisy story for our task. Therefore, we used LLMs to generate a noisy story, as shown in Figure~\ref{fig:noisy_dataset}. We first craft a prompt that transforms the detailed caption into a vibrant paragraph, brimming with emotion and imagination, while still capturing the essence of the scene. Then, we utilize the LLM to replace the adjectives in the story with their antonyms, generating a passage that contains factual errors while maintaining the key elements unchanged.

In the end, our dataset consists of paired images, noisy stories, and correct detailed descriptions. Training on this dataset can enable the model to map the text input to the corresponding image details, and then obtain the correct descriptions of these objects based on these image details.

\begin{figure}[!htp]
	\centering
	\includegraphics[width=0.9\columnwidth]{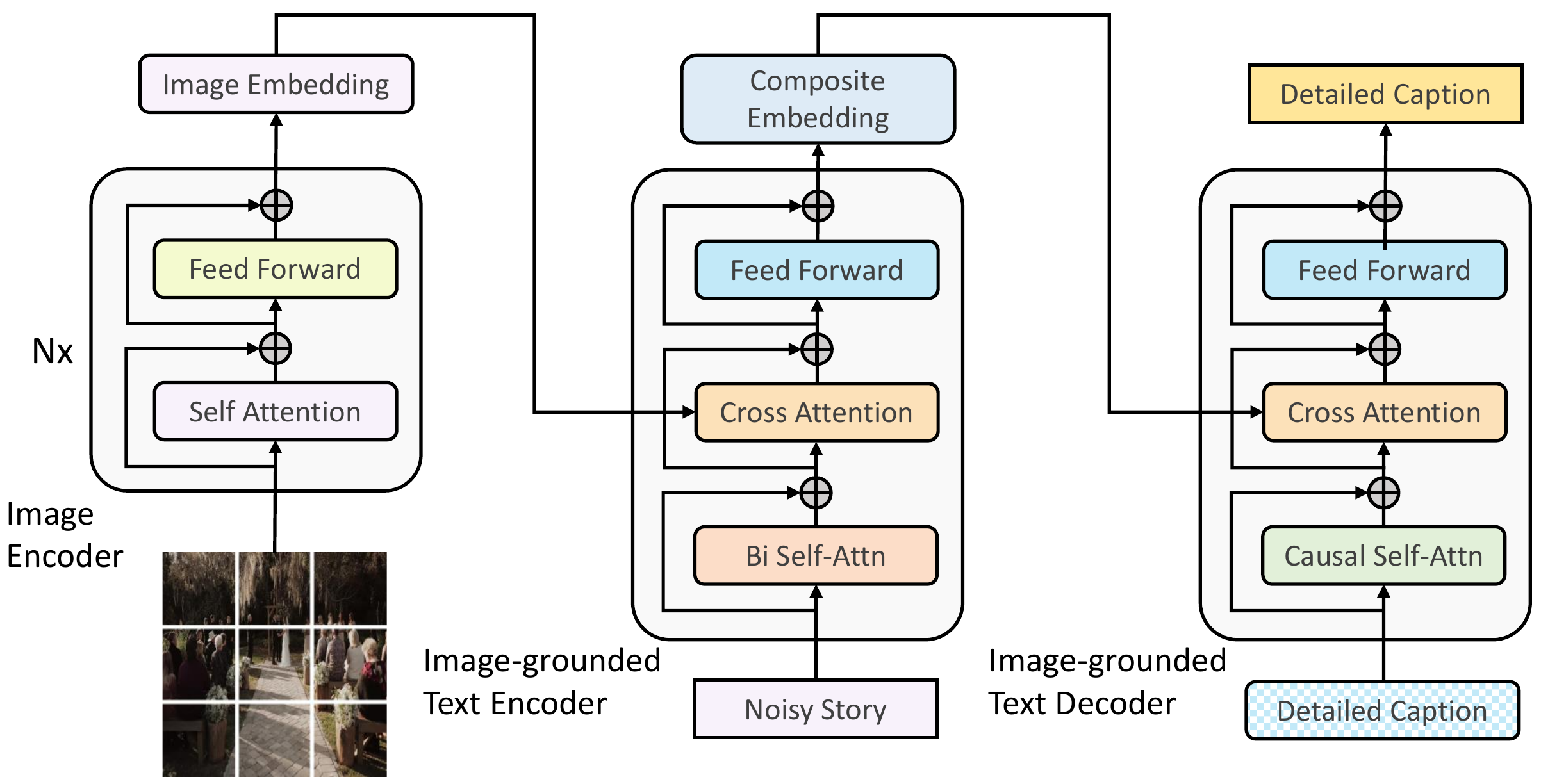}
	\caption{This figure illustrates the components of our story-aware caption model, which comprises an image encoder, a text encoder, and a text decoder. Given an input image, the encoder converts it to image embedding. Then the text encoder grounds the initiate story to the image using cross-attention and generates a composite embedding. Finally, the text decoder generates a detailed caption from the composite image-text representation.}
	\label{fig:framework}
\end{figure}

\textbf{Story-aware caption model.}
We build a story-aware caption model based on BLIP~\cite{li2022blip}, which is composed of a image encoder $g_{\text{i}}(\cdot)$, a text encoder $g_{\text{t}}(\cdot)$ and a refine decoder $d_{\text{r}}(\cdot)$, as shown in Figure~\ref{fig:framework}.

During training, the image encoder $g_{\text{i}}(\cdot)$ first transforms a image into a sequence of embedding vectors. Then, the text encoder $g_{\text{t}}(\cdot)$ takes the noisy story as input and generates a sequence of composite embedding vectors, where the cross-attentions are computed between the story embeddings and the image embeddings. Finally, the text decoder $d_{\text{r}}(\cdot)$ reconstructs the detailed caption with the composite embedding vectors inputting into the cross attentions. The model is optimized in an end-to-end way with Language Modeling loss (LM),
\begin{equation*}
    \mathcal{L}(\mathcal{U})=\sum_{i=1}^N \log \mathbb{P}\left(u_k \mid u_1, u_2,  \ldots, u_{k-1}, \Theta\right),
\end{equation*}
where $\mathcal{U}=\{u_{1}, u_{2}, \ldots, u_{N}\}$ denotes the tokens in the caption, and $\mathbb{P}\left(u_k \mid u_1, u_2,  \ldots, u_{k-1}, \Theta\right.)$ is conditional probability of $i$-th token given the previous tokens and model parameters $\Theta$.

After training, the model is capable of identifying key elements in the initial story $S^{(t)}_i$ and connecting them to corresponding regions in the image $I_i$. This information is then used to generate a more accurate and detailed description $C_i^{(t+1)}$. Using these refined captions, we propose the following prompt $p_{\text{r}}(\cdot)$ to generate more aligned stories $S^{(t+1)}_i$. In practice, we find revising the stories, rather than generating them from scratch, better preserves coherence and vividness. Below is the key part of $p_{\text{r}}(\cdot)$. The complete prompt can be found in Appendix A. 
\begin{mybox}
\textit{Given a list of json dictionaries, please use the detailed information from ``Refined Caption''  to modify the ``Initial Story'' and create a new ``Refined Story'' that better align to the real-world scenario in the photos.
}
\end{mybox}

\subsection{Iterative Refinement of Story and Image Description}

With the story-aware caption model $f(\cdot)$, we can iteratively refine the story until satisfied. 
To determine the stopping point, we introduce the concept of edit distance~\cite{levenshtein1966binary}. The iteration process ends when the ratio of the edit distance to the length of the text falls below 0.2.

However, during the iterative refinement, the stories may become overly focused on individual images, potentially losing their global perception. To address this, we propose the following prompt $p_{\text{u}}(\cdot)$ to generate a coherent and comprehensive ultimate story.

\begin{mybox}
\textit{Given a series of stories describing individual pictures from the same album, create a cohesive narrative that seamlessly connects each story together. Use appropriate transitions and scene changes to make the plot flow smoothly, while ensuring that the plot twists are logical and make sense within the context of the story.}
\end{mybox}

\section{Evaluation}
\subsection{Evaluation Dataset}
We choose to extract keyframes from popular YouTube vlogs as our primary source of images for several reasons. Firstly, video content generally exhibits higher image quality compared to individual photos found in albums. Secondly, utilizing vlogs allows us to ensure thematic consistency within a set of images, which is crucial for effective storytelling. Furthermore, YouTube offers an extensive collection of videos, and by selecting popular vlogs as our data source, we can access a wide variety of visually appealing images that are likely to resonate with a broad audience.
Our dataset comprises 30 popular YouTube videos categorized into five distinct categories: ``birthday'', ``camping'', ``christmas'', ``travel'', and ``wedding''. Each video is further divided into image collections, with each collection containing 10 key frames. The design of the dataset emphasizes the inclusion of images that possess sufficient information to support storytelling while encompassing diverse themes and styles.


Specifically, these key-frames are obtained through two steps. In the first step, meaningful key-frames are extracted and stored using FFmpeg\footnote{\url{https://www.ffmpeg.org/}}, which have high quality and often represent multiple frames within a period of time. In the second step, hand-crafted selection is performed to find a set of images that can represent the vlog. The selection criteria include a). image quality: clarity of the image; b). information content: the number of objects and elements in each image; and c). storytelling: whether these images can be strung together into a complete story, and whether they contain complete contextual relationships. The selected images range from outdoor to indoor, single to multiple scenes, and contains some dark or blur scenes to challenge the stabilisation of proposed systems.

\subsection{Automatic Evaluation Metrics}
The former VIST~\cite{huang2016visual} and VideoST~\cite{li2019video} datasets evaluate the storytelling results by comparing generated stories with given hand-craft ground truth. 
However, stories are usually too flexible to be grounded to single ground truth story, and the metric cannot measure the vividness and coherence of the stories as well.

Our systematic evaluation framework majorly evaluate two aspects of the story:
\begin{enumerate}
    \item The alignment between the stories and images;
    \item The quality of the stories.
\end{enumerate}

\textbf{EMD.} 
We propose to adopt the earth mover's distance (EMD)~\cite{rubner2000earth} to measures the distance between the distribution of the album images and the generated stories. Specifically, EMD between two distribution $P$ and $Q$ is
\begin{equation*}
    \operatorname{EMD}(P, Q)=\min _{\gamma \in \Gamma(P, Q)} \sum_{(x, y) \sim \gamma} d(x,y),
\end{equation*}
where $\Gamma(P,Q)$ is the set of all possible joint distributions of $P$ and $Q$, and $d(x,y)$ is the cost of moving unit mass from $x$ to $y$.
To compute the EMD between the images and the story, we first encode the images $\{I_i\}$ and the sentences in the story $S^{(U)} = \{T_j\}$ with the image encoder $e_{\text{i}}(\cdot)$ and text encoder $e_{\text{t}}(\cdot)$ in 
CLIP~\cite{radford2021learning} to transform the images and sentences to the same latent space, then compute the inner product as the cost function,
\begin{equation*}
    d(I_i,T_j)=\frac{e_{\text{i}}
(I_i) \cdot e_{\text{t}}(T_j)}{\left\|e_{\text{i}}(I_i)\right\| \cdot\left\|e_{\text{t}}(T_j)\right\|}.
\end{equation*}
We adopt $P$ and $Q$ as the uniform distributions on $\{I_i\}$ and $\{T_j\}$.
A lower EMD distance indicates that the generated stories are more aligned with the album images.

\textbf{LLM based evaluation metrics}.
In previous research, human evaluation was often used to measure the quality of text. However, studies~\cite{gillick2010non,clark2021all,karpinska2021perils} show that these results are not sufficiently accurate and reliable due to subjective preferences of human evaluators. On the contrary, LLMs possess extensive knowledge bases and provide more stable results, demonstrating great potential for evaluating NLP systems and algorithms~\cite{chiang2023can}. Therefore, in this article, we propose an additional evaluation metric based on LLMs, which includes the following aspects: 

\begin{itemize}
    \item Detail, which counts how many details are described in the stories. We wish the story to contain enough visual information to be aligned with the images.  
    \item Coverage, which measures the average of how much the stories  coverage the information from both short captions $\{C^{(0)}_i\}$ and detailed descriptions $\{C^{(U)}_i\}$.
    \item Coherence, which evaluates the smoothness of the stories. A good story should be logically connected, consistent, and easy to understand. 
\end{itemize}
The above metrics are implemented with GPT-4~\cite{openai2023gpt4}, the most powerful LLM so far. The complete prompts can be found in Appendix A.


\section{Result}

\subsection{Experimental Settings}
In the experiment, we utilized the GPT-3.5~\cite{ouyang2022training} as our LLM.
For the training of story-aware caption model, we adjusted the input dimensions to 480 $\times$ 480 and used a batch size of 12. The model was trained for 15 epochs using a learning rate of $2\times 10^{-5}$, which gradually decayed to 0 following a cosine learning rate schedule. The optimizer used was AdamW~\cite{loshchilov2017decoupled} with a weight decay of 0.05.
The training process was conducted on 8 Nvidia v100 32GB GPUs.

\subsection{Results of EMD distance}
We compare the performance of our proposed approach with the baseline and our proposed approach that do not use key element grounding or iterative refinement as Table~\ref{table:EMD}. The results shows that our proposed approach achieves a lower EMD distance compared to the baseline methods, indicating that our generated stories are more aligned with the album images. Moreover, our iterative approach further reduces the EMD distance, demonstrating the effectiveness of our mutually-guided approach in refining and enhancing the generated stories and image descriptions.

\begin{table}[t]
    \caption{Comparison from multi-view.}
    \label{table:EMD}
    \centering
    \setlength\tabcolsep{5.2mm}
    \renewcommand\arraystretch{0.8}
    {
    \small 
    \scalebox{1.0}{
\begin{tabular}{cccccc}
    \toprule[1pt]
    \multirow{2}{*}{Method} &\multirow{2}{*}{\#Sentence}  &\multirow{2}{*}{EMD$(\downarrow)$}  &\multicolumn{3}{c}{LLM based evaluation} \\
    \cmidrule(lr){4-6}
    & & &Detail &Coverage &Coherence\\
    \midrule 
    Captions &10.00 &12.35  &10.00 &0.85 &0.37\\
    \midrule
    Initial Story &28.70 &17.97  &40.57 &0.57 &\textbf{0.80}\\
    \midrule
    Refined Story &34.47 &16.93  &56.97 &0.60 &0.63\\
    \midrule
    Ultimate Story &34.97 &\textbf{16.23} &\textbf{60.07} &\textbf{0.62} &0.77\\
    \bottomrule[1pt]
    \end{tabular}}
    }
\end{table}

\subsection{Results of LLMs based Evaluation}

To evaluate the performance of the stories themselves, we first focused on their information content. We counted the length of sentences and the number of details included in them and found that both of these metrics increased. This indicates that with continued refinement, our framework is able to recognize more and more details from the images, resulting in more vivid and engaging stories.

The coverage metric also  increases with each step, indicating that our method is consistent with both simple global captions and detailed captions, and can align well with the real image information.

The coherence metric showed a decrease in the second step but returned to a comparable level in the third step, still higher than that of the captions. This suggests that the captions suffer from serious inconsistency issues, whereas our LLM framework generates more coherent stories using imagination, albeit with increased misalignment, as evident from the deterioration of EMD and coverage. While the second step improved the fidelity, it focused on each image and reduced coherence. Therefore, in the third refinement step, we not only improved the alignment but also enhanced the connection between independent stories, resulting in high-fidelity and highly readable stories.

\begin{figure}[!htp]
	\centering
	\includegraphics[width=1.0\columnwidth]{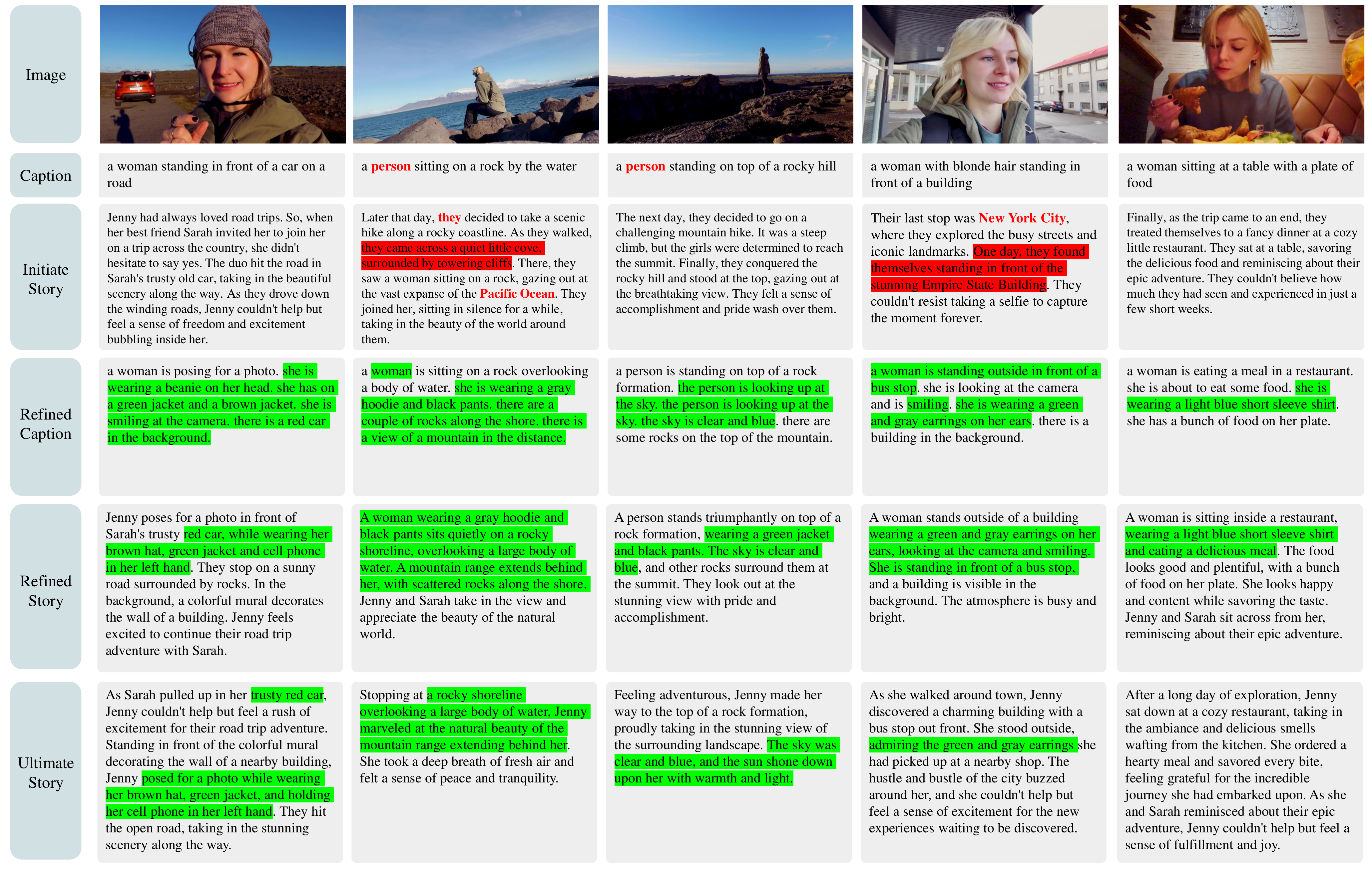}
	\caption{This figure provides a visual representation of our \texttt{VIVID}. The red sections highlight instances of unclear references or factual errors, while the green sections indicate the details that we have rectified or included. Our refined captions and stories significantly mitigate the inconsistencies between the images and texts, and the ultimate stories exhibit more coherent and cohesive.}
	\label{fig:visualization}
\end{figure}

\subsection{Visualization}
We visualize the generated stories of each step from an album in Figure~\ref{fig:visualization}. The figure shows that our \texttt{VIVID} can generate vivid stories with improved factualness during iteration.

\section{Discussion}

\textbf{Detailed caption evaluation.}
To evaluate the quality of the refined captions, we compared our proposed story-aware caption model with two baselines: directly deploying the pre-trained BLIP model~\cite{li2022blip} and fine-tuning BLIP on the \textit{Stanford image-paragraph}~\cite{krause2017hierarchical} dataset. As shown in Table~\ref{table:description}, our method outperforms both baselines on 4 common metrics: BLEU~\cite{papineni2002bleu}, ROUGE~\cite{lin2004rouge}, METEOR~\cite{banerjee2005meteor} and CIDEr\cite{vedantam2015cider}, which means our method can effectively identifying visual details.


\begin{table}[!htp]
    \caption{Result of detailed caption evaluation.}
    \label{table:description}
    \centering
    \setlength\tabcolsep{5.0mm}
    \renewcommand\arraystretch{0.8}
    {\small 
    \scalebox{1.0}{
    \begin{tabular}{cccccc}
        \toprule[1pt]
        Method &BLEU\_1 &BLEU\_4 &METEOR &ROUGE\_L &CIDEr \\
        \midrule
        BLIP~\cite{li2022blip} &0.41 &0.08 &5.42 &14.94 &0.46 \\
        \midrule
        BLIP-finetune &24.41 &7.00 &14.01 &30.86 &30.24 \\
        \midrule
        Story-aware &\textbf{51.08} &\textbf{16.42} &\textbf{22.09} &\textbf{37.93} &\textbf{72.92} \\
        \bottomrule[1pt]
        \end{tabular}}
    }
\end{table}

\textbf{Limitation and future work.} 
In Appendix B, we present the complete results for the entire dataset. While most of the results are satisfactory, there are instances where the generated stories have inconsistencies in context, misleading details, or a lack of vividness.

We recognize that fully addressing these errors through network model upgrades is challenging due to boundary effects. Therefore, we propose involving humans in the process to provide valuable assistance. In our future work, we plan to incorporate human interaction through chat to improve system performance, ultimately enhancing the practicality of our approach.



\clearpage
\bibliographystyle{plain}
\bibliography{album}


\clearpage
\setcounter{section}{0}
\renewcommand\thesection{\Alph{section}} 

\section{Details for Prompt}
\label{app:prompt}

\subsection{Detailed Prompt for Story Generation}
In this section, we provide the detailed prompt for the story generation. The results could be reproduced with the detailed prompt.

\textbf{Initial Story Generation.}
In this section, we implement a multi-step dialogue function to send the chat history and prompt to LLM, and extracts the reply, adding it to the dialogue history.

The first prompt is to generate stories from captions, and retaing the corresponding:
\begin{mybox}
    Now your answer is a set of photo captions from a vlog. Please create a vivid story that incorporates the key elements from each photo and denote the corresponding origin caption before each paragraph as "caption" "{\textbackslash}n" "generated story". Remember to use your imagination and creativity to turn the photo descriptions into a fun and engaging story. 
\end{mybox}

Then, the second prompt segment the generated stories into text chunks for the next process:
\begin{mybox}
    And please add the the generated stories into the origin json structure in the "initial$\_$story" key. For example, the answer should be like: \newline
    [\{"img$\_$path": "birthday/BpsSOqpog98/BpsSOqpog98-0190.jpg", \newline
    "caption": "a woman with glasses standing in front of a building", \newline
    "initial$\_$story": "the paragraph you generated"\}, ... ] \newline
    Please be attention, never, never, change any word in the "img$\_$path" key, because I need it to find the real photo file. And you only need to return a new json structure rather than write the real file.
\end{mybox}

\textbf{Refining the Story with Story-Aware Caption Model.}
We utilize the following prompt to revising the initial stories, with the refined captions generated by story-aware caption model:
\begin{mybox}
    Task: Given a list of json dictionaries, where each dictionary contains an "img$\_$path", a "caption", a "initial$\_$story", and a "refine$\_$caption". Please revise the "initial$\_$story" of each dictionary and store to corresponding "refine$\_$story" key, so as to better describe the real-world scenario in the "img$\_$path". The "refine$\_$caption" provides additional image information that can guide the grounding process. \newline
    Instructions: For each dictionary, use the "refine$\_$caption" to modify the "initial$\_$story" and create a new text that better describes the real-world scenario in  the "img$\_$path". Your output should be a new list of dictionaries where each  dictionary contains the original "img$\_$path" and "caption" keys, plus a new key "refine$\_$story" whose value is  the modified "initial$\_$story". Use the information from the "refine$\_$caption" to guide the grounding process and ensure that the text reflects the image information as accurately as possible. The user wants a modified text, rather than a python script. \newline
    Example Input: [\{"img$\_$path": "birthday/BpsSOqpog98/BpsSOqpog98-0136.jpg", "caption": "a basketball court surrounded by palm trees in a park", "initial$\_$story": "Carla jogged past a basketball court in the nearby park, and her mind flashed back to the beauty of the resort they had visited. She smiled as she continued her jog, grateful for the memories of that perfect day that would stay with her forever.", "refine$\_$caption": "this photo is taken outside on a sunny day. a young girl is playing in the park near a palm tree. she is wearing a pink tank top and black shorts. the park is surrounded by palm trees. the sky is blue with white clouds in it."\}, ...]   \newline
    Example Output: [\{"img$\_$path": "birthday/BpsSOqpog98/BpsSOqpog98-0136.jpg", "caption": "a basketball court surrounded by palm trees in a park", "refined$\_$story": "Carla jogged past a basketball court in the nearby park, surrounded by tall palm trees. The court was filled with young people playing basketball under the bright sun. Carla smiled as she continued her jog, enjoying the vibrant atmosphere of the park. The sky was clear and blue, with fluffy white clouds drifting lazily by. She felt grateful for the memories of that perfect day at the resort, which would always stay with her."\}, ...]   \newline
    Data: <xxx>
\end{mybox}

\textbf{Iterative Refinement of Story and Image Description.}
we propose the following prompt to generate a coherent and comprehensive ultimate story:
\begin{mybox}
    Given a series of stories describing individual pictures, with each story building upon the one before it, create a cohesive narrative that seamlessly connects each story together. Use appropriate transitions and scene changes to make the plot flow smoothly, while ensuring that the plot twists are logical and make sense within the context of the story.  \newline
    Input: <xxx>    \newline
    Tips: You should keep the number of stories. I give you 10 stories, you should return 10 stories.
\end{mybox}

\subsection{Detailed Prompt for LLM based Evaluation Metrics}
In this section, we provide the detailed prompt for the evaluation. The quality of album storytelling could be evaluated with these metrics.

\textbf{Detail.}
We counts how many details are described in the stories with the following prompt:
\begin{mybox}
    Please evaluate the input story and count the total number of details it contains. Please output the result in the format "Total number of details: xx".    \newline
    Story: <xxx>
\end{mybox}

\textbf{Coverage.}
We measures the average of how much the stories  coverage the information from both short captions $\{C^{(0)}_i\}$ and detailed descriptions $\{C^{(U)}_i\}$ with the following prompt:
\begin{mybox}
    Please use a score from 0-1 to measure how well the following story coverages the information from two different sets of captions. Note that a score closer to 1 indicates more information are covered in the story,  while a score closer to 0 indicates poorer coverage. Please output the result in the format: 
    "Score of story coverage for Caption Group 1: xx. 
    Score of story coverage for Caption Group 2: xx. 
    Average score: xx."  \newline
    Caption group 1: <xxx>  \newline
    Caption group 2: <xxx>   \newline
    Story: <xxx>
\end{mybox}

\textbf{Coherence.}
We evaluates the coherence of the stories with the following prompt:
\begin{mybox}
    Please rank the following stories on a scale of 0 to 1 based on their coherence. A score of 1 indicates that the stories are seamlessly connected and free of coherence issues, while a score of 0 indicates that there are significant coherence problems between the stories. Please consider the fact that these stories were generated independently for each image and then concatenated together. Your task is to evaluate whether there are any coherence issues between the stories when they are read together. Please output the result in the format "Coherence Score: xx".  \newline
    Story: <xxx>
\end{mybox}

\clearpage
\section{Case Study}
\label{app:case_study}
\subsection{Limited Cases}
In this section, we proved limited cases, which can be summarised into three types:
\begin{itemize}
    \item Inconsistencies, which commonly arise when the LLM encounters difficulties in comprehending the temporal sequence of storylines or establishing personal relationships.
    \item Misleading details, which means the discrete texts cannot capture all the features present in the images, or the inaccuracies in the details extracted by the story-aware caption model, resulting in erroneous stories generated by LLM. 
    \item Lack of vividness, which stems from LLM being excessively constrained by intricate details, thereby losing its creative capacity, or from the scenes being too mundane to inspire imagination.
\end{itemize}
These issued are hard to be solved by updating neural networks. In contrast, the incorporation of human interaction via chat has the potential to enhance system performance, ultimately augmenting the practicality and efficacy of our approach.

\begin{figure}[!htp]
	\centering
	\includegraphics[width=0.9\columnwidth]{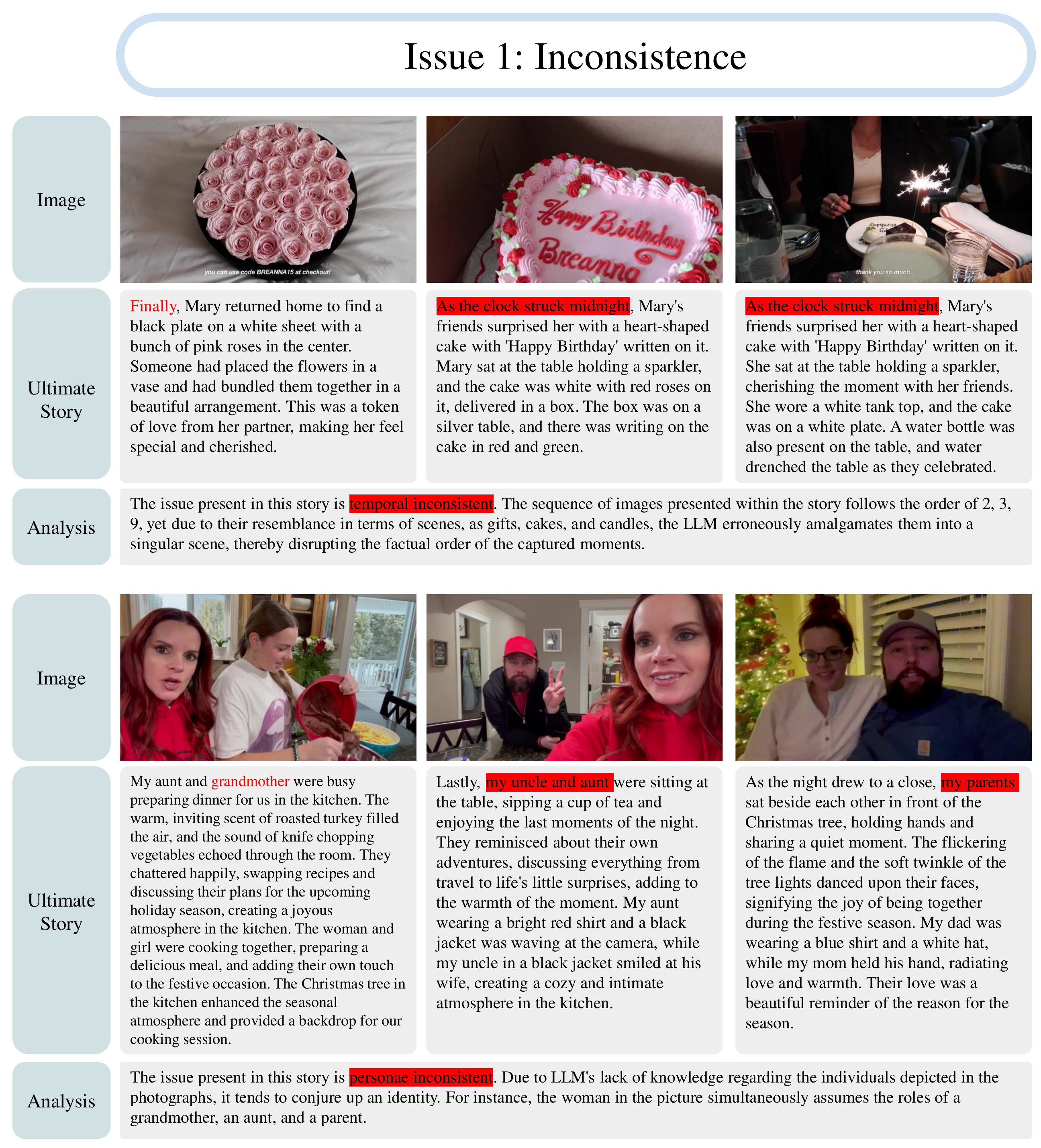}
	\caption{This figure provides samples with the ``inconsistencies'' issue.}
	\label{fig:neg_case_1}
\end{figure}

\begin{figure}[!htp]
	\centering
	\includegraphics[width=1.0\columnwidth]{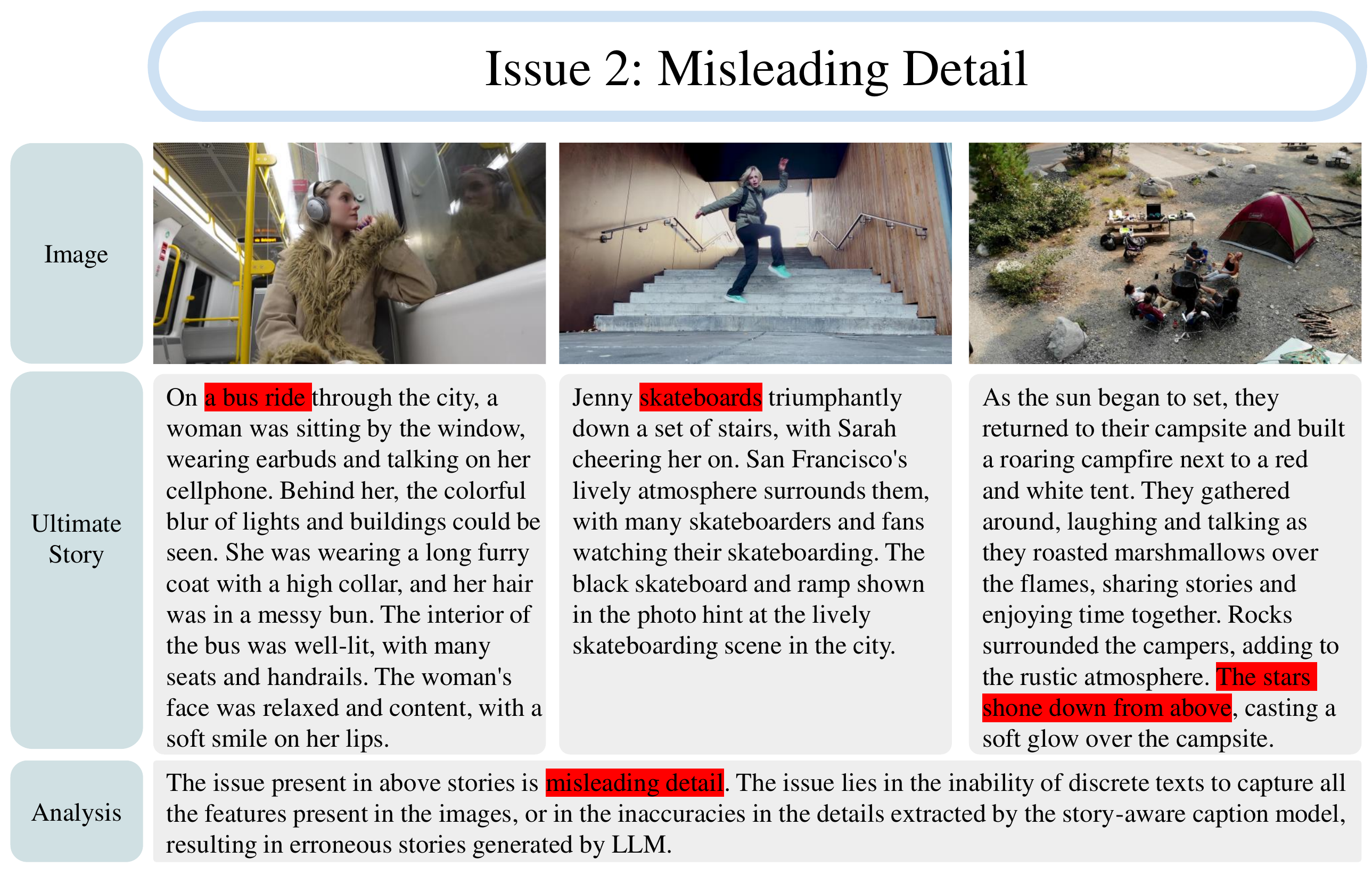}
	\caption{This figure provides samples with the ``misleading detail'' issue.}
	\label{fig:neg_case_2}
\end{figure}

\begin{figure}[!htp]
	\centering
	\includegraphics[width=1.0\columnwidth]{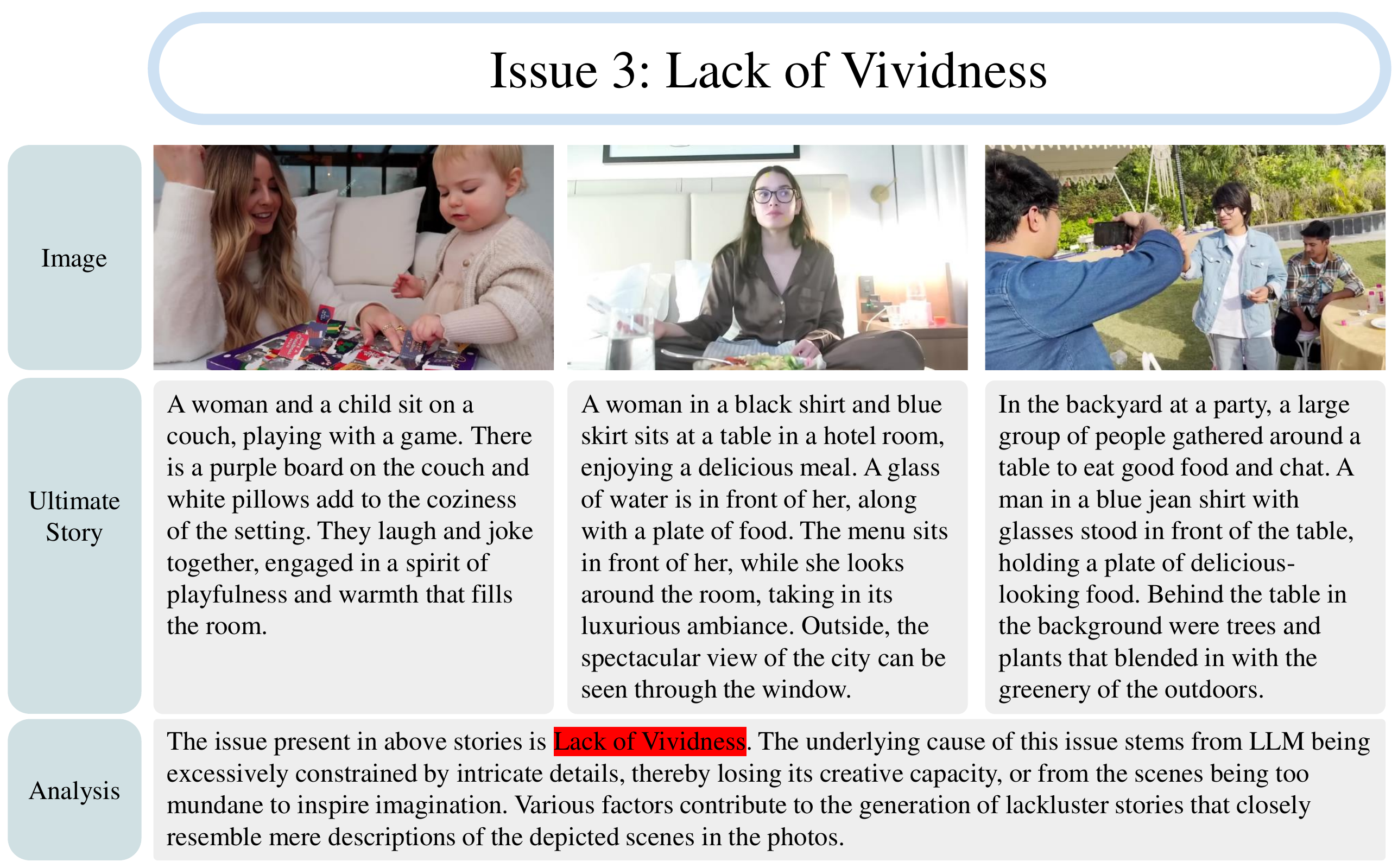}
	\caption{This figure provides samples with the ``lack of vividness'' issue.}
	\label{fig:neg_case_3}
\end{figure}

\clearpage
\subsection{Visualization of Scenes}
In this section, we present visualization samples corresponding to each scene depicted in the accompanying figures. Consistent with the main paper's settings, the red sections highlight instances of unclear references or factual errors, while the green sections indicate the details that we have rectified or included.

\begin{figure}[!htp]
	\centering
	\includegraphics[width=1.0\columnwidth]{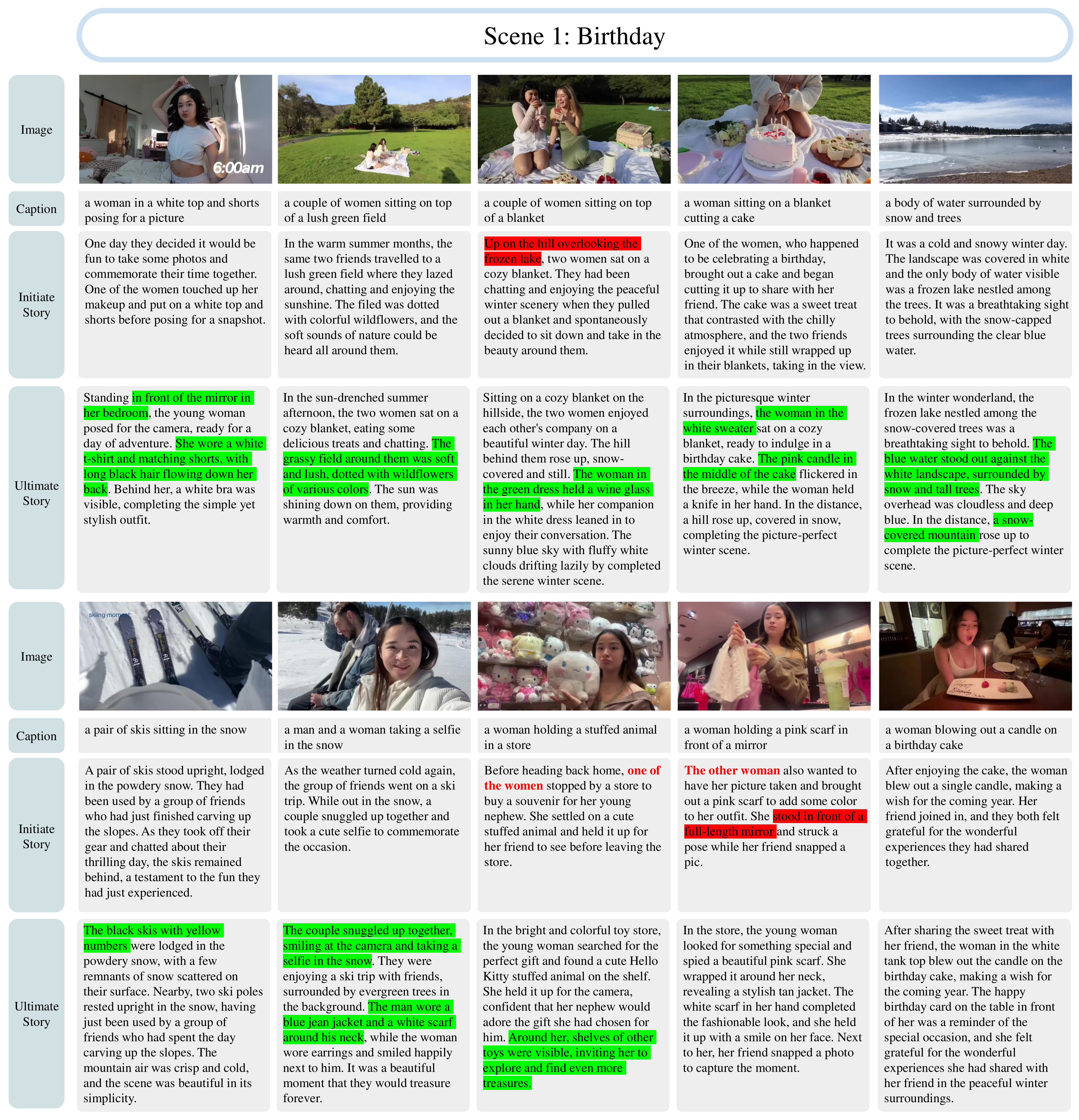}
	\caption{This figure provides an example of scene ``birthday''.}
	\label{fig:pos_case_1}
\end{figure}

\begin{figure}[!htp]
	\centering
	\includegraphics[width=1.0\columnwidth]{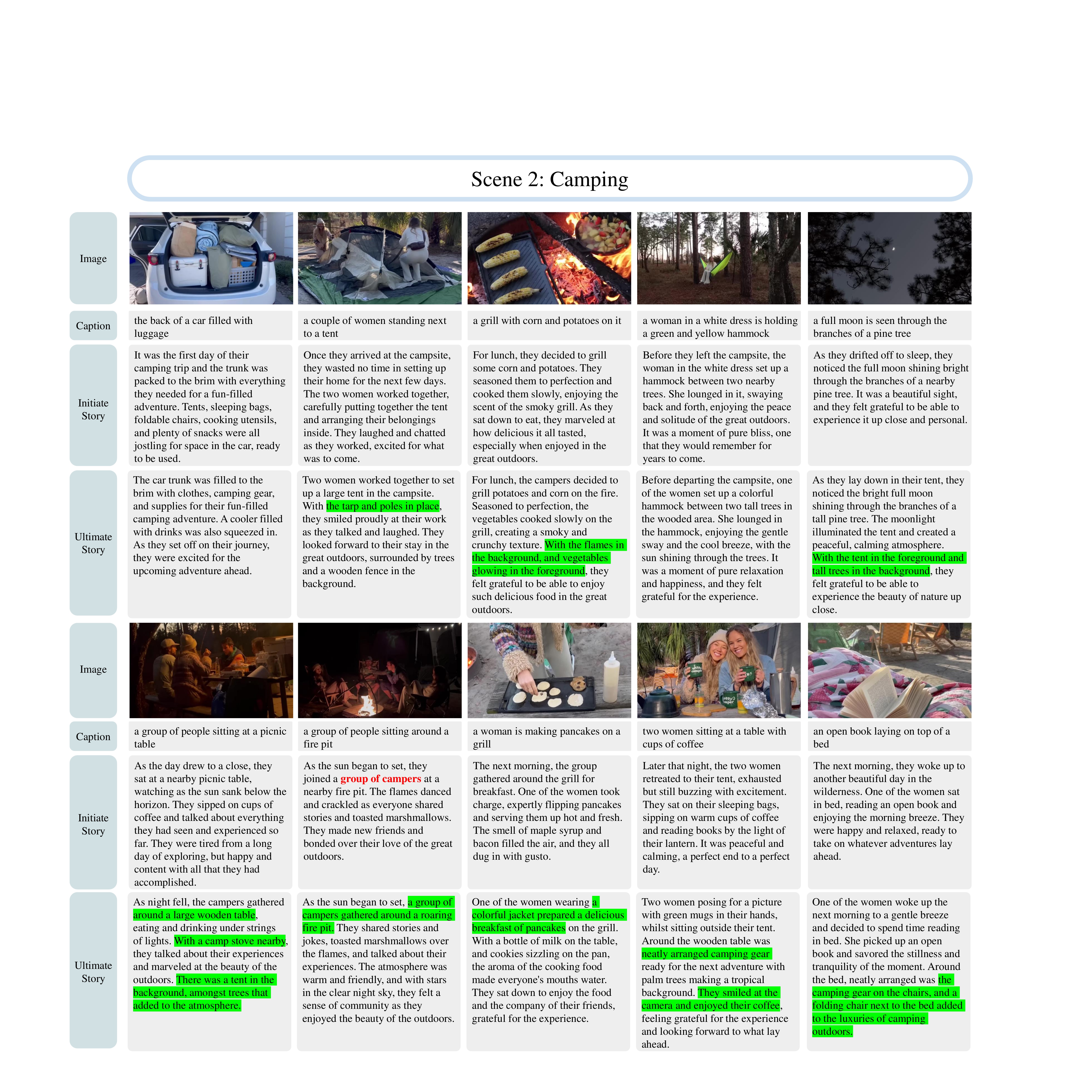}
	\caption{This figure provides an example of scene ``camping''.}
	\label{fig:pos_case_2}
\end{figure}

\begin{figure}[!htp]
	\centering
	\includegraphics[width=1.0\columnwidth]{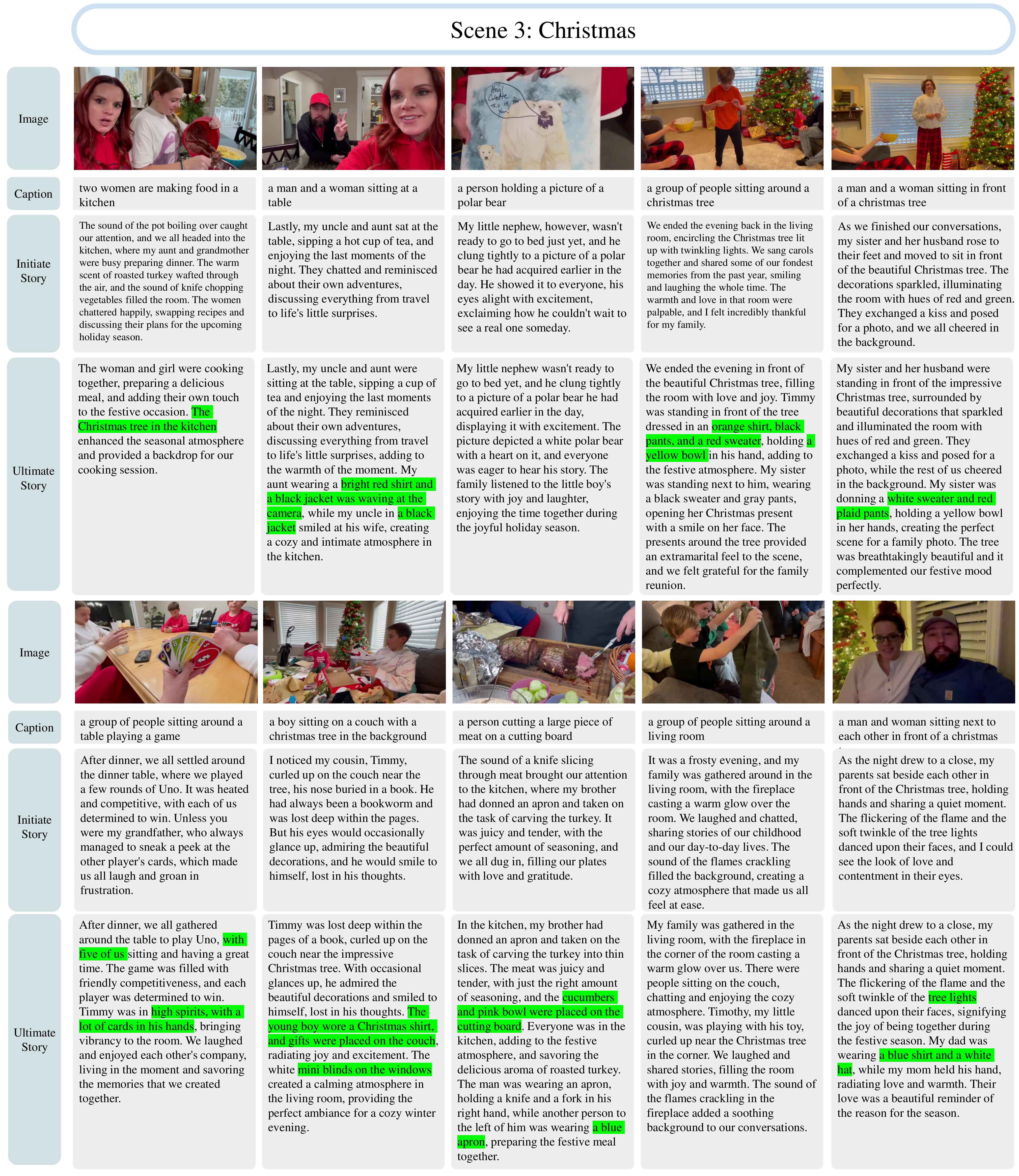}
	\caption{This figure provides an example of scene ``christmas''.}
	\label{fig:pos_case_3}
\end{figure}

\begin{figure}[!htp]
	\centering
	\includegraphics[width=1.0\columnwidth]{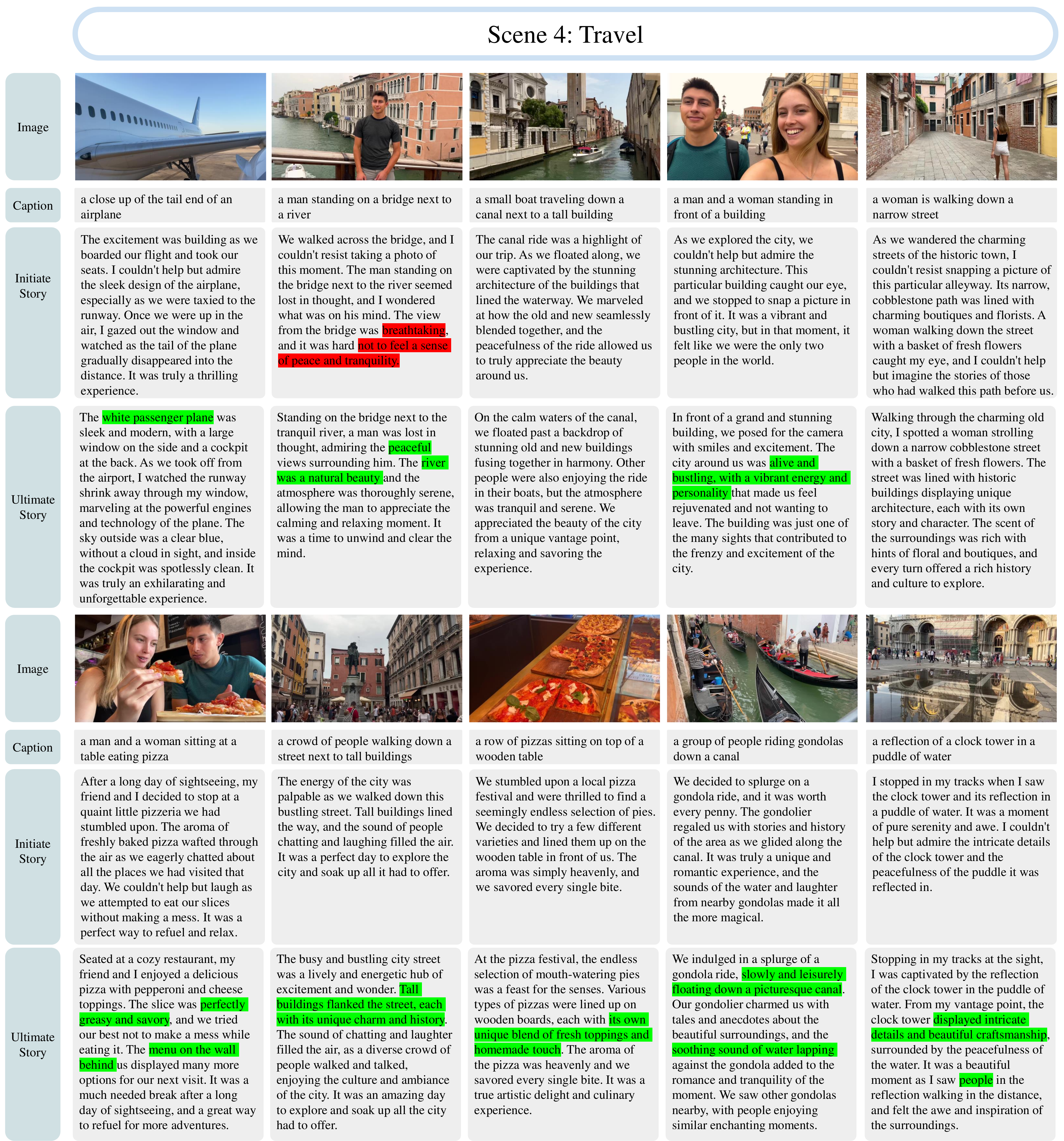}
	\caption{This figure provides an example of scene ``travel''.}
	\label{fig:pos_case_4}
\end{figure}

\begin{figure}[!htp]
	\centering
	\includegraphics[width=1.0\columnwidth]{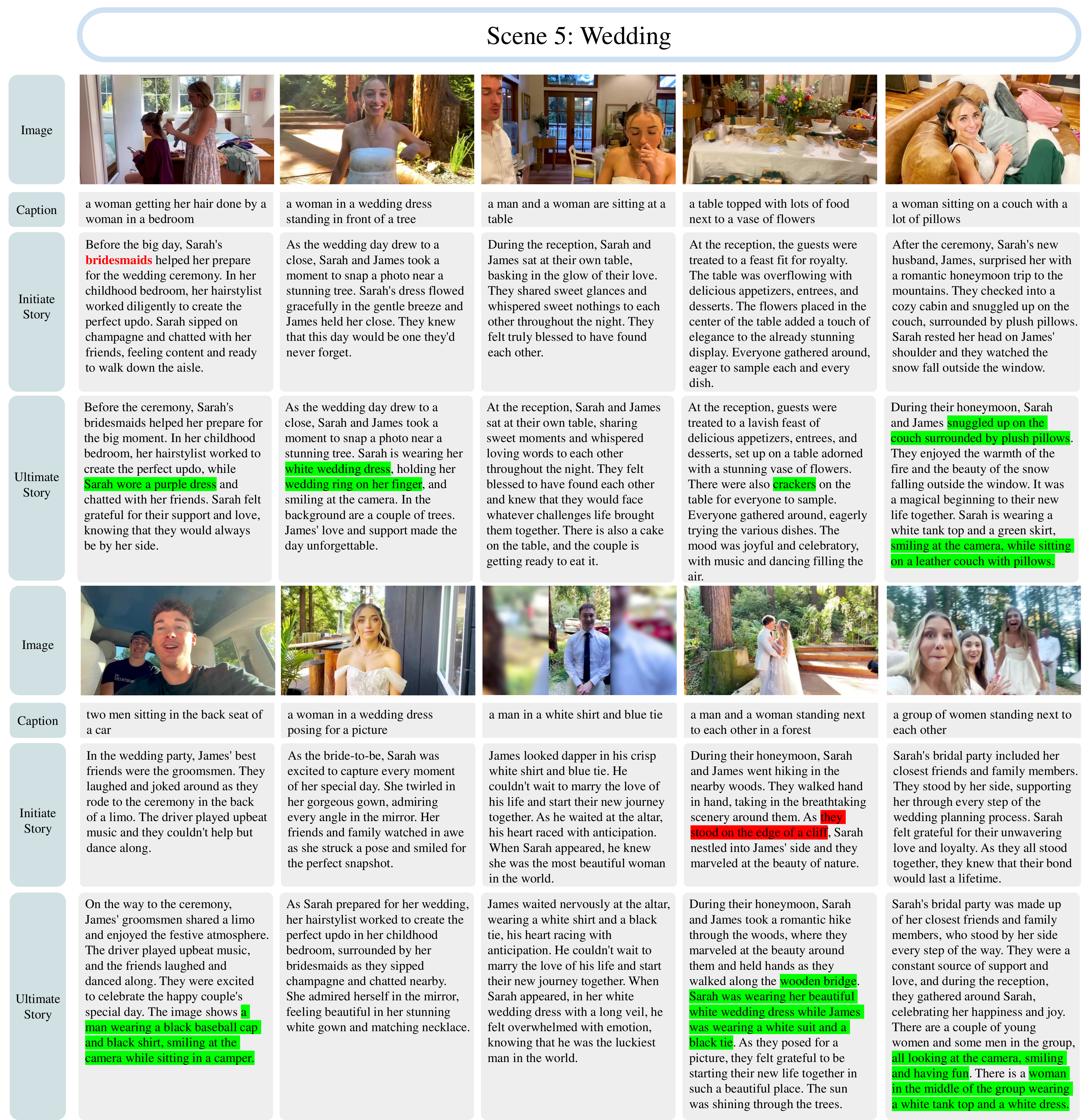}
	\caption{This figure provides an example of scene ``wedding''.}
	\label{fig:pos_case_5}
\end{figure}

\end{document}